\documentclass[12pt,a4paper]{cibb}

\usepackage{subfigure,graphicx}
\usepackage{setspace}
\usepackage{multirow}
\usepackage[usenames,dvipsnames]{xcolor}
\usepackage{amsmath,amsfonts,latexsym,amssymb,euscript,xr}

\newcommand{\ignore}[1]{}

\title{\large $\ $\\ \bf Prediction of Drug Synergy by Ensemble Learning}

\author{ I\c{s}\iıksu Ek\c{s}io\v{g}lu, Mehmet Tan}
\address{$\ $\\TOBB University of Economics and Technology\\
Department of Computer Engineering, Ankara, Turkey, \{ieksioglu, mtan\}@etu.edu.tr
}

\abstract{drug synergy, chemical representations, cancer, ensemble learning.
\\[17pt]
{\bf Abstract.} One of the promising methods for the treatment of complex diseases such as cancer is combinational therapy. Due to the combinatorial complexity, machine learning models can be useful in this field, where significant improvements have recently been achieved in determination of synergistic combinations. In this study, we investigate the effectiveness of different compound representations in predicting the drug synergy. On a large drug combination screen dataset, we first demonstrate the use of a promising representation that has not been used for this problem before, then we propose an ensemble on representation-model combinations
that outperform each of the baseline models. }

\begin{document}

\thispagestyle{myheadings}
\pagestyle{myheadings}
\markright{\tt Proceedings of CIBB 2019}

\section{\bf Scientific Background}


A drug combination is called synergistic if the effect of the drug combination on the reference cell is greater than the total effect taken from the administration of the individual drugs. If the opposite situation is observed, the drug combination is called antagonistic \ignore{\cite{jia2009mechanisms}}. Understanding whether a combination is antagonistic or synergistic is a resource and time intensive task. Developing new synergistic combinations or finding the best disease-specific combination therapy is also a complex problem, because high dimensional search space requires searching correct parameters. 



Recently, machine learning methods have made significant contributions to the studies of combination therapies \ignore{\cite{tsigelny2018artificial}}.  In this article, we compare three different drug representation methods to see how they effect predictions of synergy scores and how predictions change according to these representations and learning models. These representations are the characteristic direction \cite{clark2014characteristic} vectors inferred from drug-induced gene expression profiles, a chemical fingerprint representation \cite{preuer2017deepsynergy} and drug vectors which are learned directly from molecular graphs by using a graph neural network \cite{tsubaki2018compound}. We used the target synergy scores as given in \cite{o2016unbiased} and \cite{preuer2017deepsynergy}. In addition, we propose an ensemble method that outperforms each of the individual methods involved in predicting the synergy score. 

There are a number of recent studies for predicting the effect of drug combinations by using machine learning. \ignore{Huang and colleagues \cite{huang2014systematic} tried to predict synergistic drug combinations with side effects. In their work, instances are modeled using side effects as features. Drug combinations were labeled according to these side effects and useful drug combinations were selected.} NLSS \cite{chen2016nllss} is a work that uses semi-supervised learning to discover new synergistic drug combinations. Assumption of \cite{chen2016nllss} is that, the more similar the drugs are to each other, the greater the synergistic effect.  Ensemble methods were also previously used to predict drug synergy score \cite{singh2018prediction}. The authors employed several well known machine learning algorithms, among which they pick the best performing four and combined predictions with specific weights to predict actual synergy score estimates. Unlike our study, they worked only on a single representation of drugs.  Another well known ensemble learning method, extremely randomized trees, were applied to this problem in  \cite{jeon2018silico}. Features used in this study are drug-target information, gene expression and mutation datasets, pharmacological data and synthetic lethality.   DeepSynergy \cite{preuer2017deepsynergy} is a fully connected neural network to predict drug synergy. It includes one input layer, two hidden layers and one output layer. Concatenated drug and cell line features are given as input. Promising results were obtained with a fully connected architecture. On the same dataset,  \cite{janizek2018explainable} applies extreme gradient boosting algorithm. They observed that this tree based model outperforms the neural networks. In addition, this type of method is more explainable than a neural network. 

None of above mentioned studies have explored the effect of drug representations for different algorithms. The ensemble methods mentioned above only consider different methods not different representations. To our best, this is the first study that investigates prediction of drug synergy on this context.  

The rest of the article is organized as follows. Descriptions of the methods and data are given in Section \ref{sec2}. Section \ref{sec3} reports the setting and experimental results. Finally, we conclude in Section \ref{conc} and give some future research directions.

\section{\bf Materials and Methods}
\label{sec2}

Drug and cell line representations in addition to the methods employed are explained in this section. First, we describe the methods for drug synergy score prediction and then shortly explain the different representations used. The synergy scores were taken from a large compound oncology dataset~\cite{o2016unbiased} which is used for training the model in~\cite{preuer2017deepsynergy} where the deviation from a theoretical model of synergy (Loewe additivity \ignore{\cite{Loewe1953}} in this case) was calculated and used as the drug synergy score.


\subsection{Methods}
\label{methods}

We used four different single types of models to predict drug synergy scores. These are classical neural networks, elastic nets, gradient boosted decision trees, random forests and graph neural networks. Short descriptions of these models follow.


A fully connected neural network (FCNN) is one of the most basic neural network types. 
It is the type that is composed of layers where 
each neuron in layer $k$ is connected to every node in layer $k+1$. The weights of the network is optimized by minimizing a given error measure through backpropagation algorithm. 
Note that this is actually the same model proposed in \cite{preuer2017deepsynergy}.


Graph embedding is a group of methods to map the graph data to a low dimensional space \ignore{\cite{cai2018comprehensive}}. 
Embedding can be performed for nodes, edges or the whole graph. In this paper, to construct one of the representations
for drugs, we employ whole graph embedding and use Graph Neural Networks (GNN). GNN is the name given to a group of 
neural network methods that can directly operate on a graph input. The specific GNN we use is the architecture proposed in \cite{tsubaki2018compound}. We modified this architecture to work with drug pairs and to estimate the synergy score. First, graph molecules(i.e., r-radius sub-graphs' atom types) in each pair  are embedded separately in d-length vectors with random weights. Then, initialized vectors are passed through two separate FCNNs with L layers. Initialized vectors are updated according to graphs' adjacency matrices when passing through the L-layered FCNN. Output vectors of the last layers of the L-layered FCNN are the representations that will be used for prediction. These embeddings combined with the cell line descriptors can now be used with any learning model that predicts drug synergy. 



Elastic net is a regularized linear regression method that uses both L1 and L2 regularization in the target function. Therefore, $\beta_i$ values optimizing, $L(\alpha,\beta) =  (y - \hat{f}(x))^2 + (1 - \alpha) \sum_{i=1}^{n} \beta_i^2 + \alpha \sum_{i=1}^{n} |\beta_i|$ constitute the model, where $\beta_i$'s are the parameters of the linear regression model $\hat{f}$. 



A Gradient boosting machine (GBM) is an additive model where a weak learner is fitted to the current residuals at each iteration. At the end, all 
the predictions from the weak learners scaled by a small learning rate are added to compute the final prediction.
In this paper, we used decision trees as the weak learners to train a GBM.



Random Forest (RF) \ignore{\cite{ho1995random}} is an ensemble algorithm that combines many decision trees with bootstrap aggregation (bagging) technique \ignore{\cite{breiman1996bagging}} and feature sampling. The estimation of a new instance is performed by computing the average of the responses of all models. 

In addition to and using the above methods, we propose an ensemble method composed of "drug representation-method" combinations as base learners. We used a fixed representation for the cell lines (the gene expression profiles) to correctly observe the effect of drug representations. Among other ensemble learning methods such as stacked generalization with non-linear learners as the meta learner, weighted linear combination of the base learners performed best. This method is different from the previously proposed methods for this problem in the sense that each learner in the ensemble may use different representations of drugs. The results demonstrate the effectiveness of the proposed ensemble methodology.



\subsection{Cell line and drug representations}

In this paper, to represent cell lines, we used the genomics features used in~\cite{preuer2017deepsynergy}. These features were obtained by applying Factor Analysis for Robust Microarray Summarization (FARMS) \ignore{\cite{hochreiter2006new}} to the quantile normalized cell line gene expression profiles downloaded from ArrayExpress database (E-MTAB-3610). At the end of this process, a vector of length 3984 were obtained as the descriptor of cell lines.

There are three types of drug representations we use. First one is composed of some of the well known chemical descriptors. Second one employs the drug-induced gene expression profiles of cell lines and therefore represents a drug by using it's effects on gene expression. Finally the last one is a graph neural network method that only uses the graph topology of drugs to form a representation vector.


The descriptors that we name as the chemical descriptors are the ones used in~\cite{preuer2017deepsynergy}. In this representation a compound is described as a vector of length 4387 composed of 1309 extended connectivity fingerprint with radius 6 (ECFP\_6) features \ignore{\cite{rogers2010extended}}, 802 physico-chemical features obtained by ChemoPy \ignore{\cite{cao2013chemopy}}and 2276 toxicophore features obtained from the liteature. We will denote this representation as ChemR.

Identification of differentially expressed genes (DEG) is a basic task in gene expression analysis. 
Characteristic direction (CD) \cite{clark2014characteristic} is a 
geometric approach to determine DEG. In \cite{wang2016drug} CD is computed to identify DEG between 
normal and drug-induced gene expression profiles from Library of Integrated Cellular Signatures (LINCS) 
\cite{subramanian2017next} project. This CD vector for each drug is of length 978 that comprises of the DEG status of 
landmark genes identified by LINCS. In this paper, as the second representation for drugs, we employ the CD computed 
from the normal and drug-induced gene expression data. The data is available from 
 \cite{wang2016drug}. We will denote this representation as CDR.

The third drug representation is denoted GNNR, learned by using the GNN with the structure described in Section \ref{methods}. After this model has been trained, d-length GNNR representation of any drug can be obtained by using L-layer FCNN which is part of the architecture of the GNN we used. Since the L-layer FCNN is dependent on the GNN, the vectors that this structure maps are actually the output of an intermediate layer of GNN. Thus, the output vectors of this intermediate layer constitutes the GNNR representation. These vectors, which are created to minimize the error of the FCNN model, are recorded and used as a representation for compounds.

Using three different representations and four different models mentioned above, we investigated which representation-model combinations perform better for drug synergy prediction. In order to accomplish this, first we identified the instances that are common in all three datasets. Only 29 of the 38 drugs that have ChemR representations are available within CDR. Instances that do not have CDR data are not included in our study.
Thus, the number of samples we use is 24780 corresponding to these common drugs. Each of these instances are composed of two drugs and a cell line with a corresponding synergy score. To compensate the ordering of drugs, the first half of the 24780 feature vectors are of type drugA-drugB-cell line, where the other half consists of the same instances with drugs re-ordered (drugB-drugA-cell line), as in \cite{preuer2017deepsynergy}. Except GNNR, tanh normalization is applied to CDR and ChemR similar to \cite{preuer2017deepsynergy}. 

\section{\bf Experimental Results}
\label{sec3}

Representation-model combinations were compared according to the scores obtained as a result of 5-fold cross validation. Cross validation was performed by dividing datasets with the "leave drug combinations out" method, as in \cite{preuer2017deepsynergy}.

We performed parameter optimization for the models on a single validation split. 
Hyperparameter tuning search space for gradient boosting and elastic net are selected according to \cite{janizek2018explainable}. For all ensemble tree models, number of estimators is selected as 1000. For CDR and ChemR datasets, in random forest and gradient boosting, maximum tree depths $M \in \{6,8\}$, also in gradient boosting the learning rate $T \in \{0.05,0.1\}$. For GNNR, $M \in \{2,4\}$, $T \in \{0.05,0.01\}$. For Elastic Net, in all datasets, $\alpha \in \{0.25,0.5,0.75,1.0,1.15,1.25,1.5\}$. For FCNN, there are five different hyperparameters we optimized; number of nodes in two hidden layers ($H$), learning rate
 ($F$), dropout rate ($D$) and number of epochs ($E$). There are also three GNN specific parameters, drug representation dimension ($d$), radius ($r$) and the number of layers of L-layered FCNN ($L$). For ChemR, $H \in \{\{8192,4096\},\{4096,2048\}\}$, $F \in \{0.0001,0.00001\}$. For CDR, $H \in \{\{2500,3000,3500\}, \{1000,1500,2000,2500\}\}$, $D \in \{0.33,0.4\}$ and $E \in \{350, 500,455,200\}$. For GNNR, $H \in \{\{2000,3000,4000\}, \{1000,1500,2000\}\}$, $E \in \{500,1000\}$, $d \in \{25,50,100\}$, $r \in \{2,3\}$ and $L \in \{3,4\}$.  Scikit-learn \ignore{\cite{scikit-learn}} is employed for Elastic Net and Random Forest models, keras \ignore{\cite{chollet2015keras} }and pytorch \ignore{\cite{ketkar2017introduction}} are used for GNN and FCNN models. We utilized xgboost library \ignore{\cite{Chen:2016:XST:2939672.2939785}} for gradient boosting.


\begingroup
\setlength{\tabcolsep}{5pt}
\renewcommand{\arraystretch}{0.725}
\begin{table}[t]
\vspace{0.5mm}
\caption{Results for all drug representation-model combinations and proposed ensemble model\label{data1}}
\begin{center}
\begin{tabular}{| *{9}{p{0.07\linewidth}|} }
    \hline
  & \multicolumn{4}{c|}{$\scriptscriptstyle MSE$}
            & \multicolumn{4}{c|}{$\scriptscriptstyle Pearson$}               \\
    \hline
  & $ \scriptscriptstyle FCNN$    &   $ \scriptscriptstyle GB$  &   $ \scriptscriptstyle RF$  &   $ \scriptscriptstyle Elas. N.$  &   $ \scriptscriptstyle FCNN$  &   $ \scriptscriptstyle GB$  &   $ \scriptscriptstyle RF$ &   $ \scriptscriptstyle Elas. N.$    \\
    \hline
\multicolumn{1}{|l|}{$\scriptscriptstyle CDR$ }  &\multicolumn{1}{|l|}{$\scriptscriptstyle266.0 \pm 57.9$}&\multicolumn{1}{|l|}{$\scriptscriptstyle295.8 \pm 61.3$}    &\multicolumn{1}{|l|}{$\scriptscriptstyle405.1 \pm 76.6$ }    &\multicolumn{1}{|l|}{$\scriptscriptstyle451.4 \pm 76.6$  }   &\multicolumn{1}{|l|}{$\scriptscriptstyle0.74 \pm 0.04$ }    &\multicolumn{1}{|l|}{$\scriptscriptstyle0.69 \pm 0.03$ }    & \multicolumn{1}{|l|}{$\scriptscriptstyle0.56 \pm 0.03$}    &\multicolumn{1}{|l|}{ $\scriptscriptstyle0.47 \pm 0.03$ }  \\
    \hline
\multicolumn{1}{|l|}{$\scriptscriptstyle ChemR$}   &\multicolumn{1}{|l|}{$\scriptscriptstyle273.7 \pm 53.7$}       &\multicolumn{1}{|l|}{$\scriptscriptstyle 295.2 \pm 55.9$}       &\multicolumn{1}{|l|}{$\scriptscriptstyle410.9 \pm 63.5$}       & \multicolumn{1}{|l|}{$\scriptscriptstyle452.0 \pm 77.4$}      &\multicolumn{1}{|l|}{$\scriptscriptstyle0.72 \pm 0.03$}       &\multicolumn{1}{|l|}{$\scriptscriptstyle0.7 \pm 0.03$}       &\multicolumn{1}{|l|}{$\scriptscriptstyle0.54 \pm 0.05$}     &\multicolumn{1}{|l|}{$\scriptscriptstyle0.47 \pm 0.03$}   \\
    \hline
\multicolumn{1}{|l|}{$\scriptscriptstyle GNNR$ }  & \multicolumn{1}{|l|}{$\scriptscriptstyle306.4 \pm 55.9$ }     &\multicolumn{1}{|l|}{$\scriptscriptstyle572.5 \pm 105.9$ }      &\multicolumn{1}{|l|}{$\scriptscriptstyle578.2 \pm 101.8$ }      &\multicolumn{1}{|l|}{$\scriptscriptstyle583.4 \pm 103.8$}       & \multicolumn{1}{|l|}{$\scriptscriptstyle0.69 \pm 0.03$}      &\multicolumn{1}{|l|}{$\scriptscriptstyle0.15 \pm 0.01$ }      &\multicolumn{1}{|l|}{$\scriptscriptstyle0.14 \pm 0.01$}      &\multicolumn{1}{|l|}{$\scriptscriptstyle0.11 \pm 0.02$}   \\
    \hline
\multicolumn{1}{|c|}{$\scriptscriptstyle Ensemble$}     & \multicolumn{4}{|c|}{$\scriptscriptstyle 260.112 \pm 57.144$} & \multicolumn{4}{|c|}{$\scriptscriptstyle 0.745 \pm 0.035$}   \\
  \hline
\end{tabular}
    \end{center}
\end{table}
\endgroup

We can denote each representation-model combination as $r^m$ where $m$ is the model trained on representation $r$.
Parameters that give the best result for $CDR^{FCNN}$ are $E=455, H=\{3000,1500\}$ and $D=0.4$. The best performing parameters for $CDR^{GB}$ are $M=6$ and $T=0.05$ as in \cite{janizek2018explainable}. The best parameter combination for $ChemR^{FCNN}$ is the $\alpha=0.00001, H=\{8192, 4096\}$. For $ChemR^{GB}$, the best parameters are $M=6$ and $T=0.05$. For $GNNR^{FCNN}$  (i.e., $GNN$), the best performing parameters are $E=1000, r=2, d=25, L=3$ and $H=\{3000,1500\}$. These are the parameters selected for the five best representation-model combinations they are the ones used in our ensemble model. The best parameters for others are not given due to space restrictions.

To construct the ensemble, we followed a greedy forward selection procedure. The models, first, are sorted in decreasing order by their mean squared error (MSE). Then, starting from the top, each model is added to the ensemble one-by-one. With each new model, we searched for the optimal weights (which are in $[0,1]$) of each model in the ensemble by a grid search procedure. We stopped adding models to the ensemble when we observe a decrease in performance. This procedure resulted in five representation-model combinations in the final ensemble, where these models are $CDR^{FCNN}$, $ChemR^{FCNN}$, $CDR^{GB}$, $GNNR^{FCNN}$ and $ChemR^{GB}$ with weights $\{0.535, 0.19, 0.15, 0.065, 0.06\}$ respectively.

The results are given in Table \ref{data1} in terms of MSE and Pearson correlation ($\rho$). It is observed that the proposed ensemble significantly outperforms the other representation-model ($r^m$) combinations ($p<0.05$ for all pairwise comparisons by Wilcoxon signed rank test). FCNN consistently outperforms all other single models both in terms of MSE and $\rho$ on all representations. Then comes the GBM, RF and
elastic net respectively. This gives us a hint that the relationship here is a non-linear one. Our findings 
here are also inline with \cite{preuer2017deepsynergy} that a neural network model seems to be suitable for this problem. Also the proposed ensemble model outperforms the DeepSynergy model (given as $ChemR^{FCNN}$ in our results) proposed in \cite{preuer2017deepsynergy}. 

For representations, the winner is not as clear as the models. In terms of average MSE for all models, CDR is better
than others, while there is no difference between CDR and ChemR for $\rho$. GNNR performs poor 
especially for GBM, RF and elastic net, only $GNNR^{FCNN}$ is close to others in terms of performance. 
This is expected as the GNN model only uses the graph topology as input compared to others that also 
include chemical and genomics descriptors. Also, if we look at the individual representation-model combinations ($r^m$) $CDR^{GB}$ and $ChemR^{GB}$ are not significantly different from each other ($p=0.753$). The 
situation is also similar for $CDR^{FCNN}$ and $ChemR^{FCNN}$ ($p=0.115$).


\begin{figure}[h]
\vspace{3mm}
 \begin{center}
 \includegraphics[width=14cm]{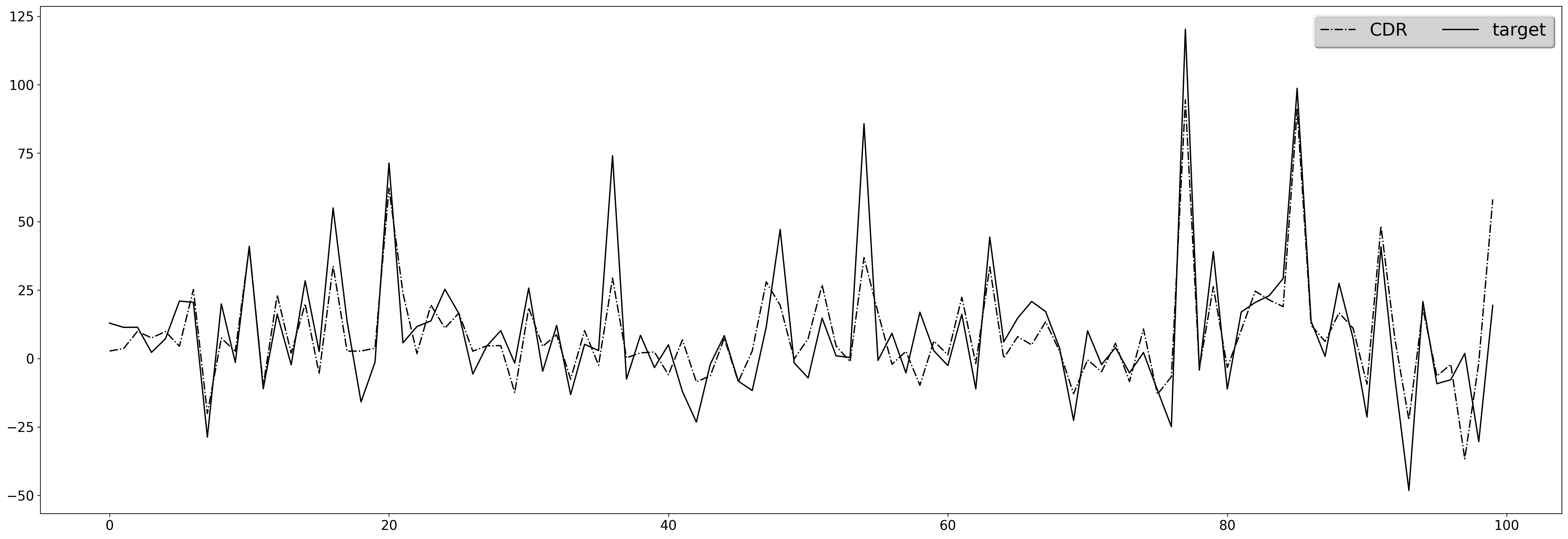}
\caption {difference between the targets and the estimates obtained using CDR data of 100 samples\label{cibb-fig1}}
 \end{center}
\vspace{-8mm}
\end{figure}

\begin{figure}[h]
\vspace{3mm}
 \begin{center}
 \includegraphics[width=14cm]{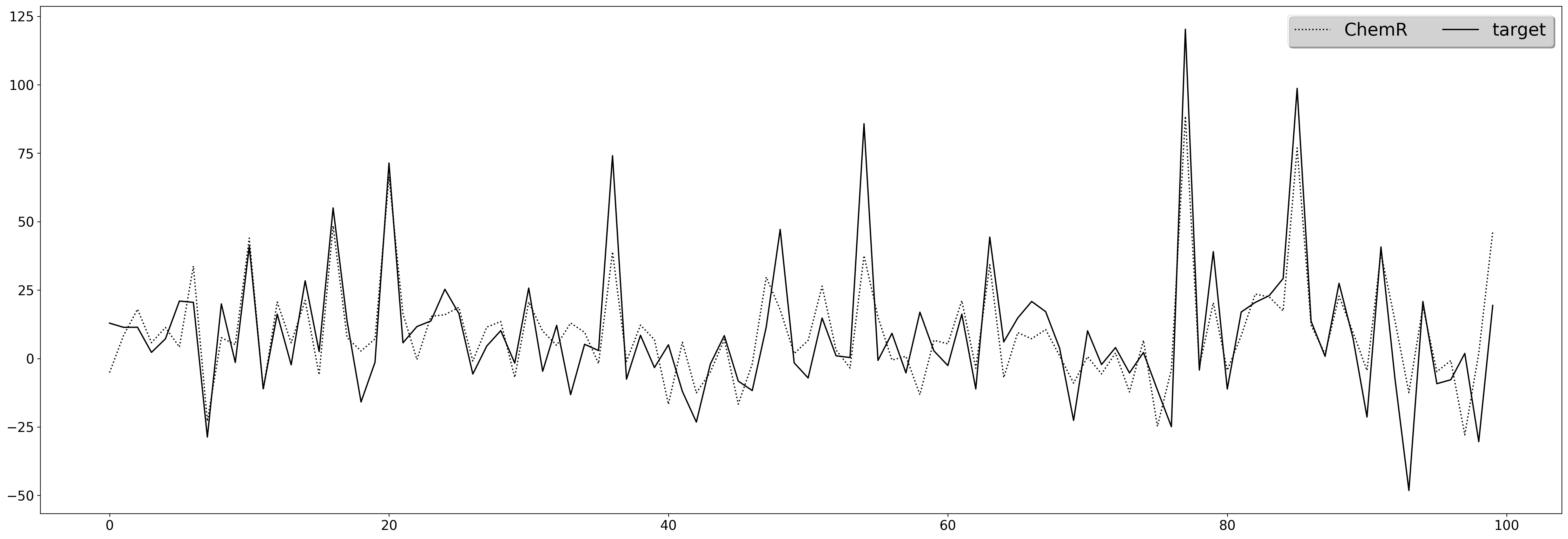}
\caption {difference between the targets and the estimates obtained using ChemR data of 100 samples\label{cibb-fig2}}
 \end{center}
\vspace{-8mm}
\end{figure}

\begin{figure}[h]
\vspace{3mm}
 \begin{center}
 \includegraphics[width=14cm]{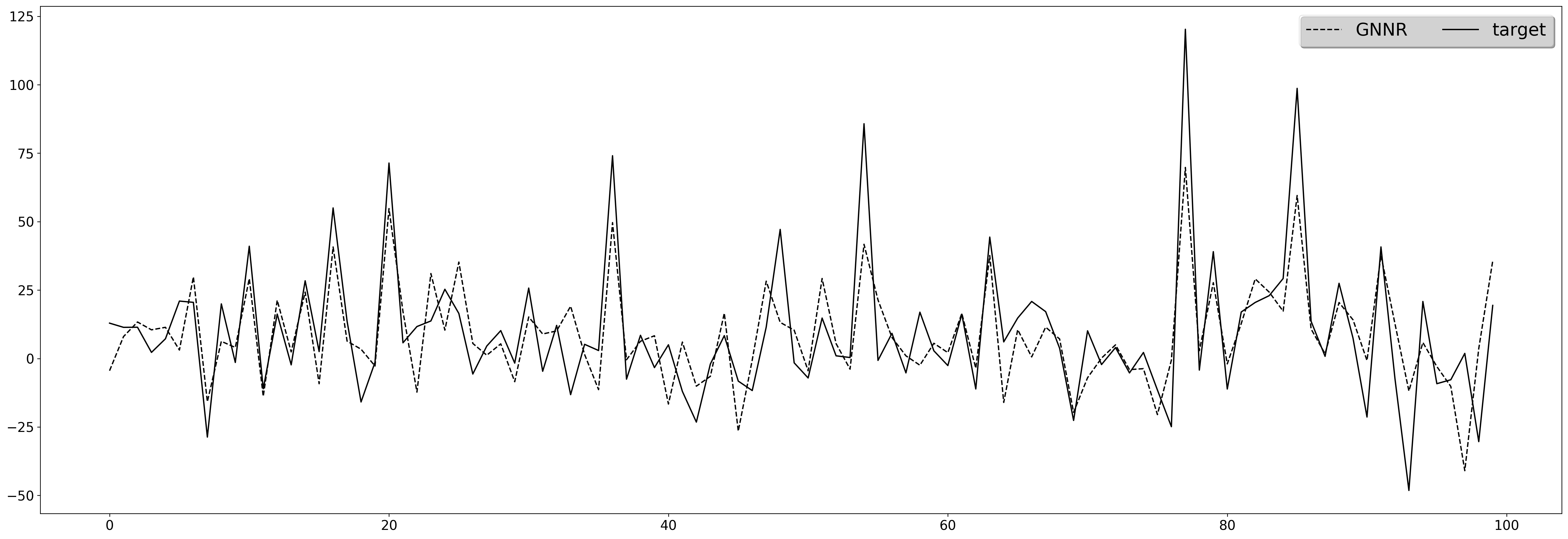}
\caption {difference between the targets and the estimates obtained using GNNR data of 100 samples\label{cibb-fig3}}
 \end{center}
\vspace{-8mm}
\end{figure}

Figures \ref{cibb-fig1},\ref{cibb-fig2} and \ref{cibb-fig3} show a comparison between the estimations obtained using different representations with FCNN and actual values of 100 randomly selected samples. It can be seen that the results obtained with CDR and ChemR are very similar to each other where the predictions obtained from both datasets generally achieved the capturing distribution of target values. On the other hand, it can be observed that the experiment with GNNR dataset failed to predict the extreme points. For this representation, in general, estimated values are close to the average of the target values.

To sum up, CDR is a promising representation in terms of performance and computational complexity. However, 
it may not be available for any compound, as it's availability is limited by the LINCS database. In that case, 
one may use the ChemR representation. GNNR is also promising in the sense that it performs close to other models by only using the graph topology. Finally, it is clear that using both representation and method ensembles helps
improving the prediction of drug synergy.

\section{\bf Conclusion}
\label{conc}


In this paper, we investigated the effect of compound representations on predicting drug synergy. We employed three different representation types on four different single models. We demonstrated that a representation summarizing the drug-induced gene expression profiles may help improving the performance. Also we proposed an ensemble model
that can perform better than each of the baselines, demonstrating the usefulness of ensemble learning for this type of problems. There are a few directions that we plan to extend this work. First, we plan to extend the number of model and representation types. Second, we will work identifying the features that are more
important in each of the representations. This may lead to better understanding the mechanisms underlying drug synergy. 



\bibliographystyle{apalike}
{\fontsize{10}{10}\selectfont

}

\end{document}